\documentclass[10pt,twocolumn,letterpaper]{article}

\usepackage{cvpr}
\usepackage{times}
\usepackage{epsfig}
\usepackage{amsmath}
\usepackage{amssymb}
\usepackage{graphicx}
\usepackage{subcaption}
\usepackage{multirow}
\usepackage{enumerate}

\newcommand{\comment}[1]{}

\usepackage[pagebackref=true,breaklinks=true,letterpaper=true,colorlinks,bookmarks=false]{hyperref}

\cvprfinalcopy 

\def\cvprPaperID{0346} 

\ifcvprfinal\fi
\begin{document}

\title{The 1st Agriculture-Vision Challenge: Methods and Results}

\author{{\small
Mang Tik Chiu$^{1*}$, Xingqian Xu$^{1}$\thanks{~indicates joint first author. For more information on our database and other related efforts in Agriculture-Vision, please visit our CVPR 2020 workshop and challenge website {\color{blue} https://www.agriculture-vision.com}.}~~, Kai Wang$^{3}$, Jennifer Hobbs$^2$, Naira Hovakimyan$^{2,1}$, Thomas S. Huang$^1$, Honghui Shi$^{3,1}$}\\
{\small Yunchao Wei$^{1}$, Zilong Huang$^{1}$, Alexander Schwing$^{1}$, Robert Brunner$^{1}$,Ivan Dozier$^2$,  Wyatt Dozier$^2$, Karen Ghandilyan$^2$,}\\
{\small David Wilson$^2$, Hyunseong Park$^4$, Junhee Kim$^{4,5}$, Sungho Kim$^{4}$, Qinghui Liu$^6$, Michael C. Kampffmeyer$^7$, Robert Jenssen$^7$,}\\
{\small Arnt B. Salberg$^6$, Alexandre Barbosa$^1$, Rodrigo Trevisan$^1$, Bingchen Zhao$^8$, Shaozuo Yu$^8$, Siwei Yang$^8$, Yin Wang$^8$, Hao Sheng$^9$,}\\
{\small  Xiao Chen$^9$, Jingyi Su$^{10}$, Ram Rajagopal$^9$, Andrew Ng$^9$, Van Thong Huynh$^{11}$, Soo-Hyung Kim$^{11}$, In-Seop Na$^{12}$}\\
{\small Ujjwal Baid$^{13}$, Shubham Innani$^{13}$, Prasad Dutande$^{13}$, Bhakti Baheti$^{13}$, Sanjay Talbar$^{13}$, Jianyu Tang$^{14}$}\\
\\
{\small $^1$UIUC, $^2$Intelinair, $^3$University of Oregon, $^4$Agency for Defense Development, South Korea,}\\
{\small $^{5}$DGIST, South Korea, $^6$Norwegian Computing Center, $^7$UiT The Arctic University of Norway,} \\
{\small $^8$Tongji University, China, $^9$Stanford University, $^{10}$Chegg, Inc., $^{11}$Chonnam National University, South Korea,}\\
{\small $^{12}$Chosun University, South Korea, $^{13}$SGGS Institute of Engineering and Technology, India, $^{14}$Tsinghua University, China}}

\maketitle


\begin{abstract}
    The first Agriculture-Vision Challenge aims to encourage research in developing novel and effective algorithms for agricultural pattern recognition from aerial images, especially for the \textbf{semantic segmentation} task associated with our challenge dataset. Around 57 participating teams from various countries compete to achieve state-of-the-art in aerial agriculture semantic segmentation. The \textbf{Agriculture-Vision Challenge Dataset} was employed, which comprises of 21,061 aerial and multi-spectral farmland images. This paper provides a summary of notable methods and results in the challenge. Our submission server and leaderboard will continue to open for researchers that are interested in this challenge dataset and task; the link can be found \href{https://competitions.codalab.org/competitions/23732?secret_key=dba10d3a-a676-4c44-9acf-b45dc92c5fcf}{here}.
\vspace{-10mm}
\end{abstract}

\section{Introduction}
Vision in agriculture has begun gaining increasing attention as recent advancements in deep learning solutions for various tasks were proven successful. Areas such as medicine and aerospace~\cite{larson2017performance,aniyan2017classifying,yu2019novel, yu2017computed,yu2020foal} have benefited from the effectiveness of vision applications in their respective domains. As a result, there have been numerous efforts that aim to apply pattern recognition techniques in agriculture to increase potential yield as well as prevent losses. Nevertheless, progress in these directions have been slow~\cite{kamilaris2018deep}, which can be partially attributed to the lack of datasets that encourage relevant studies.

Semantic segmentation from aerial agricultural images, as one of the major topics in agriculture-vision applications, differs from common object or aerial image segmentation tasks in several aspects. First, farmland images are usually multi-spectral, since image channels such as near-infrared and thermal inputs are extremely helpful for field anomaly detection. Second, different from common objects with clear boundaries, farmland patterns are regions with extremely irregular shapes and scales. These distinctions make aerial agricultural image semantic segmentation a uniquely challenging task with great academic and economic potentials.

Nevertheless, inspirations for agricultural semantic segmentation can be drawn from methods aimed for common object segmentation. Recent works on segmentation in general have demonstrated impressive results~\cite{wei2018revisiting, cheng2019spgnet,huang2019ccnet, qian2019weakly, jiao2019geometry, huang2020alignseg, wang2020alleviating, wang2020differential}.
For example, SPGNet~\cite{cheng2019spgnet} leverages multi-scale context modules to improve semantic segmentation performances. The DeepLab series~\cite{chen2014semantic,chen2016deeplab,chen2017rethinking,chen2018encoder} uses atrous convolution to further expand the receptive field, which enhanced the network's ability to capture objects at larger scales. CCNet~\cite{huang2019ccnet} proposed a criss-cross convolution to more efficiently capture non-local features. These techniques can potentially be transferred to semantic segmentation in agricultural images to yield similar performance gains.

Motivated by the above, the first Agriculture-Vision Challenge was held to encourage research in this area. A subset of the original Agriculture-Vision dataset~\cite{chiu2020agriculture} (i.e. the Agriculture-Vision Challenge dataset) was used. The challenge dataset contains 21,061 aerial and multi-spectral farmland images captured throughout 2019 across the US. In the following sections we describe and discuss in detail the challenge, notable methods and results.
\newcommand{\img}[1]{
    \includegraphics[width=\linewidth]{IMAGES/crop_vis/#1.jpg}
}

\newcommand{\imgbox}[2]{
    \begin{subfigure}{#1\linewidth}
    \centering
    #2
    \end{subfigure}
}

\begin{figure*}[t!]
\centering
\vspace{0.1cm}
    \imgbox{0.32}{
        \imgbox{0.9}{\img{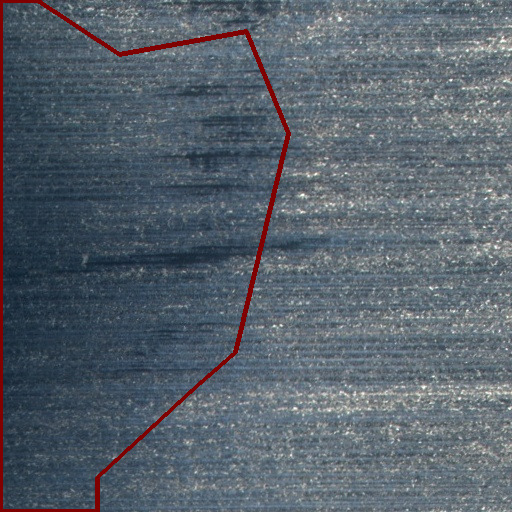}}%
        \vspace{-0.5cm}
        \caption{Cloud shadow}
        \vspace{0.5cm}
    }%
    \imgbox{0.32}{
        \imgbox{0.9}{\img{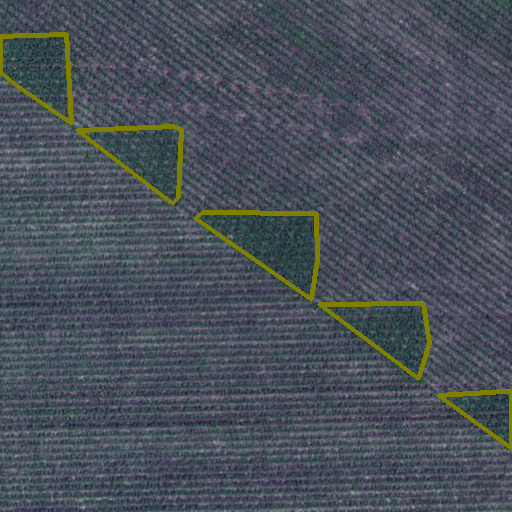}}%
        \vspace{-0.5cm}
        \caption{Double plant}
        \vspace{0.5cm}
    }%
    \imgbox{0.32}{
        \imgbox{0.9}{\img{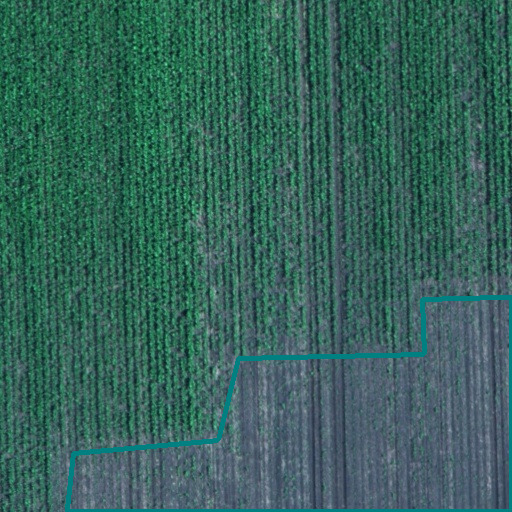}}%
        \vspace{-0.5cm}
        \caption{Planter skip}
        \vspace{0.5cm}
    }
    \imgbox{0.32}{
        \imgbox{0.9}{\img{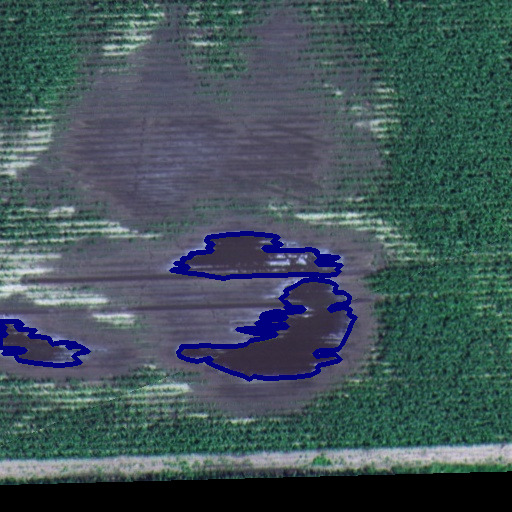}}%
        \vspace{-0.5cm}
        \caption{Standing water}
        \vspace{0.5cm}
    }%
    \imgbox{0.32}{
        \imgbox{0.9}{\img{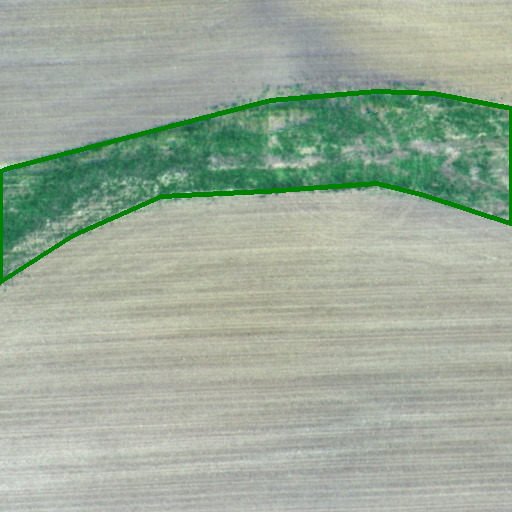}}%
        \vspace{-0.5cm}
        \caption{Waterway}
        \vspace{0.5cm}
    }%
    \imgbox{0.32}{
        \imgbox{0.9}{\img{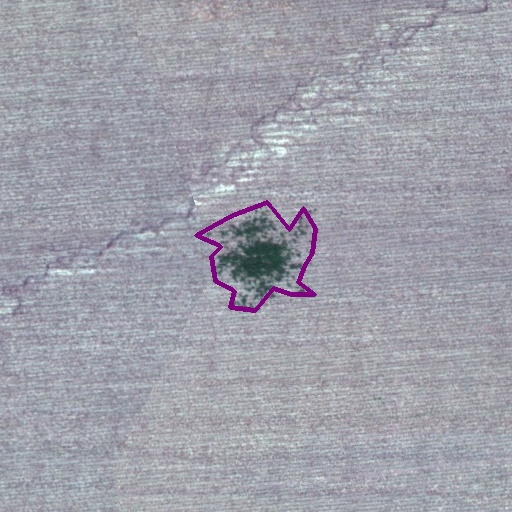}}%
        \vspace{-0.5cm}
        \caption{Weed cluster}
        \vspace{0.5cm}
}
\caption{RGB images of each pattern in the challenge dataset. Note that as the original Agriculture-Vision dataset~\cite{chiu2020agriculture} is updated, more patterns are gradually being included. Images best viewed with color and zoomed in.}
\label{fig:crop_vis}
\end{figure*}

\section{The Agriculture-Vision Challenge}
\subsection{Challenge Dataset}
The first Agriculture-Vision Challenge focuses on semantic segmentation from aerial agricultural images. Six important anomaly patterns from the Agriculture-Vision dataset~\cite{chiu2020agriculture} are to be recognized, which are cloud shadow, double plant, planter skip, standing water, waterway and weed cluster. Each image is $512\times512$ pixel with four input channels, namely Red, Green, Blue and Near-infrared (NIR). In addtion to the input channels, a boundary map and a mask are provided to indicate areas within the farmland and the valid pixels in the image respectively. In total, the challenge dataset contains 12901/4431/3729 train/val/test images respectively. Visualization of each pattern is shown in Figure~\ref{fig:crop_vis}. Note that labels in this dataset are not mutually exclusive, which means that a pixel can contain more than one pattern. As a result, a custom metric is designed to evaluate submissions.

\subsection{Evaluation Metric}
To accommodate for overlapping labels, we modify the conventional mean Intersection-over-Union (mIoU) metric by categorizing predictions of any label in a pixel as a correct prediction. This enables easy adaptation of typical semantic segmentation models into our agriculture challenge.

Specifically, to compute the modified mIoU, a confusion matrix $M^{c\times c}$ ($c=7$ is the number of classes plus background) is first computed with the following rules:\\

For each prediction $x$ and label set $Y$ at a pixel:
\begin{enumerate}[\indent\indent(1)]
    \item If $x\subseteq Y$, then $M_{y,y}=M_{y,y}+1\quad\forall y\in Y$
    \item Otherwise, $M_{x,y}=M_{x,y}+1\quad\forall y\in Y$
\end{enumerate}

Finally, the modified mIoU is computed by:
\begin{align*}
    \frac{1}{c}\sum_c\frac{True\  positive_c}{Prediction_c + Target_c - True\ positive_c}
\end{align*}

The modified mIoU increases the reward for a correct prediction by allowing any correct predictions to count as true positives for all ground truth labels. However, it also heavily penalizes the model if the prediction does not match any of the ground truth labels.

\begin{table*}[t!]
    \centering
    \begin{tabular}{r||c||c|c|c|c|c|c|c}
    \multirow{2}{*}{Submission}&modified&Back-&Cloud&Double&Planter&Standing &Water-&Weed\\
    &mIoU&ground&shadow&plant&skip&water&way&cluster\\\hline\hline
    DSSC&63.9&80.6&56&57.9&57.5&75&63.7&56.9\\
    seungjae&62.2&79.3&44.4&60.4&65.9&76.9&55.4&53.2\\
    yjl9122&61.5&80.1&53.7&46.1&48.6&76.8&71.5&53.6\\
    SCG\_Vision&60.8&80.5&51&58.6&49.8&72&59.8&53.8\\
AGR&60.5&80.2&43.8&57.5&51.6&75.3&66.2&49.2\\
SYDu&59.5&81.3&41.6&50.3&43.4&73.2&71.7&55.2\\
agri&59.2&78.2&55.8&42.9&42&77.5&64.7&53.2\\
TJU&57.4&79.9&36.6&54.8&41.4&69.8&66.9&52\\
celery030&55.4&79.1&38.9&43.3&41.2&73&61.5&50.5\\
stevenwudi&55&77.4&42&54.4&20.1&69.5&67.7&53.8\\
PAII&55&79.9&38.6&47.6&26.2&74.6&62.1&55.7\\
agrichallenge12&54.6&80.9&50.9&39.3&29.2&73.4&57.8&50.5\\
hui&54&80.2&41.6&46.4&20.8&72.8&64.8&51.4\\
shenchen616&53.7&79.4&36.7&56.3&21.6&67&61.8&52.8\\
NTU&53.6&79.8&41.4&49.4&13.5&73.3&61.8&56\\
tpys&53&81.1&50.5&37.1&25.9&67.4&58.7&50.1\\
Simple&52.7&80.2&40&45.2&24.6&70.9&57.6&50.4\\
Ursus&52.3&78.9&36.3&37.8&34.4&69.3&57.1&52.3\\
liepieshov&52.1&77.2&40.2&46&16&71.3&62.9&51.1\\
Lunhao&49.4&79.5&40.4&38.8&10.5&69.4&58.3&49.1\\
tetelias-mipt&49.2&80.4&37.8&34.8&4.6&70.6&62.5&53.8\\
Dataloader&48.9&79.1&42&35.8&9.1&68.7&56.7&51.3\\
Hakjin&46.4&78.6&32&38.3&1.8&66.2&58&49.9\\
JianyuTANG&44.6&78.1&37.9&31.8&15.4&47.3&54.8&46.9\\
Haossr&43.9&79.2&21.4&28.1&2.7&67.5&56.4&52.3\\
rpartsey&41.5&72.5&21.6&36.2&9.1&59.7&40.7&50.6\\
TeamTiger&40.8&75.2&26.1&40.1&9.9&48&37.1&49.5\\
Chaturlal&40.7&77.7&23&20.4&5&55&51&52.9\\
Sciforce&40.2&80.5&29.6&24.4&0&41.2&55.9&50\\
MustafaA&40.1&76.5&34.4&25.6&11.1&46&36.5&50.3\\
HaotianYan&36.8&77.1&21.9&25.1&13.7&57.5&24.3&37.9\\
gro&36.3&76.4&37.5&8.4&0&60.3&29.7&41.8\\
oscmansan&35.5&71.6&29.6&3&0&52.4&46.2&45.9\\
ThorstenC&33.6&72.3&22.3&10&2&40.8&40.1&47.8\\
ZHwang&33.5&76.5&32.4&12.9&0&57.2&15.9&39.9\\
fayzur20&22.1&65.4&21.8&2.2&0.2&23.3&13.4&28.7\\
gaslen2&21.5&71&3.3&17.9&0.8&10.2&6.9&40.1\\
dvkhandelwal&16.3&71.5&0&0&0&42.6&0&0\\
ajeetsinghiitd&10.3&56.9&0.2&0.4&0&0&0.1&14.5\\
    \end{tabular}
    \caption{Challenge results ranked by modifeid mIoU.}
    \label{tab:results}
\end{table*}

\subsection{Challenge Description}
The first Agriculture-Vision challenge was hosted between January 27, 2020 and April 20, 2020. Around 57 teams participated in the challenge, with about 33 publicized result submissions. Submissions were evaluated on the challenge test set with 3729 images and ranked based on the modified mIoU.

\section{Results and Methods}
Table~\ref{tab:results} shows the results of the first Agriculture-Vision challenge. In this section, we review some notable submission, such as their motivations and methodologies.

\subsection{Team DSSC}
\vspace{-2mm}
\noindent\textbf{Hyunseong Park, Junhee Kim, Sungho Kim}\\
\noindent Agency for Defense Development, DGIST
\vspace{1mm}

\begin{figure}[h!]
    \centering
    \includegraphics[width=\linewidth]{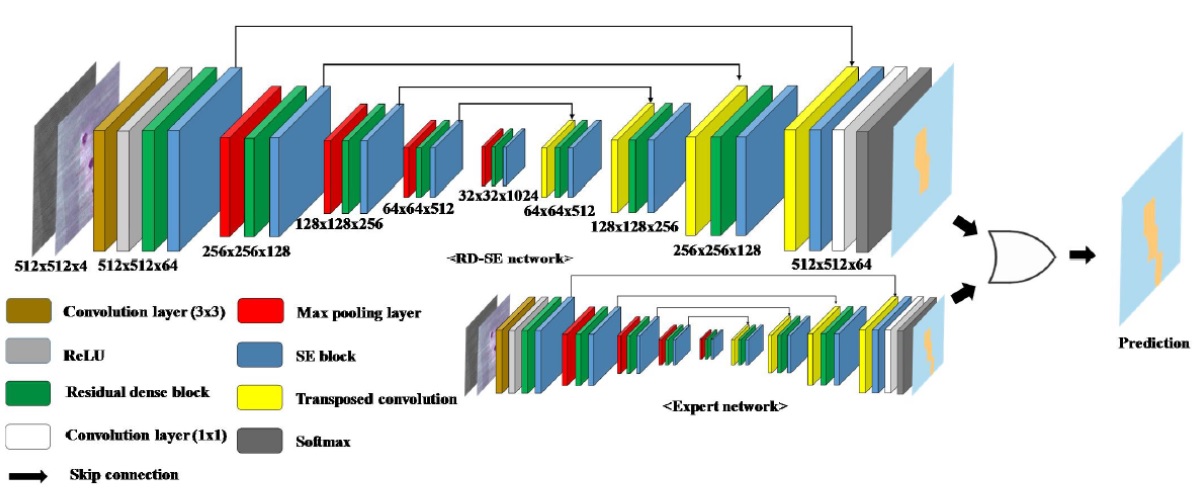}
    \caption{Team DSSC: Residual DenseNet with Expert Network architecture.}
    \label{fig:rd_se}
\end{figure}

Residual DenseNet~\cite{zhang2018residual} with Squeeze-and-Excitation blocks~\cite{hu2018squeeze} (RD-SE) is adopted as the base model for semantic segmentation. RD-SE is based on U-Net~\cite{ronneberger2015u} architecture that has encoder/decoder architecture as shown in Figure~\ref{fig:rd_se}. In RD-SE, to compensate for the spatial loss which arise during the feature extraction, residual dense blocks~\cite{zhang2018residual} and skip connections are utilized. Also, Squeeze-and-Excitation blocks (SE block)~\cite{hu2018squeeze} are used to recalibrate channel-wise feature responses. Five convolution layers with kernel size 3x3 and batch normalization~\cite{ioffe2017batch} are included in one residual dense block.

Expert networks were also used to segment less frequent class objects. In this challenge, two expert networks are trained for minor classes (i.e. planter skip and standing water). The planter skip expert network takes also the double plant images as training input since many planter skip patterns also appear in the same image with double plant patterns. Therefore, the planter skip expert network considers 3 classes (i.e. planter skip, double plant and background). Although expert networks are based on RD-SE, they have a lighter architecture and can be trained faster than RD-SE. Trained expert networks support RD-SE to segment minority patterns. The overall process is shown in Figure~\ref{fig:rd_se}. From the input images, RD-SE networks produce the prediction maps. If there are pixels classified as planter skip, the expert networks are implemented to segment on the same images. The prediction results for both RD-SE and expert networks are combined to make final prediction. Unlike expert networks for planter skip, the expert networks for standing water is used when there are pixels classified as planter skip and standing water from RD-SE. The result requires several steps of post-processing, including transition from planter skip to standing water (when both labels appear in the same field), removal of small labels and morphological closing.

\subsection{Team SCG\_Vision}
\vspace{-2mm}
\noindent\textbf{Qinghui Liu, Michael C. Kampffmeyer, Robert Jenssen, Arnt B. Salberg}\\
\noindent Norwegian Computing Center, UiT The Arctic University of Norway
\vspace{1mm}

\begin{figure}[h!]
    \centering
    \includegraphics[width=\linewidth]{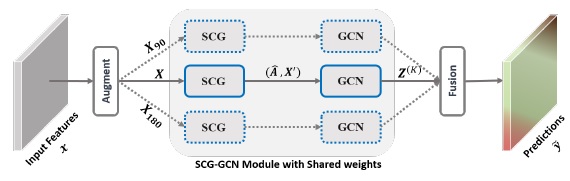}
    \caption{Team SCG\_Vision: Multi-view Self-Constructing Graph Convolutional Network architecture.}
    \label{fig:scg_gcn}
\end{figure}

The proposed model uses the self-constructing graph (SCG)~\cite{liu2020self} module combined with graph convolutional network~\cite{kipf2016semi} for aerial agricultural semantic segmentation. Since aerial images are rotational invariant, three SCG-GCN modules are used to extract features at multiple views. The proposed model architecture is shown in Figure~\ref{fig:scg_gcn}.

To overcome the class imbalance problem in the challenge dataset, an adaptive class reweighing loss is designed. A positive-negative class balanced function is further adopted to accommodate for negative samples. Details of this work can be found in our workshop proceedings: \textbf{Multi-view Self-Constructing Graph Convolutional Networks with Adaptive Class Weighting Loss for Semantic Segmentation}.

\subsection{Team AGR}
\vspace{-2mm}
\noindent\textbf{Alexandre Barbosa, Rodrigo Trevisan}\\
\noindent University of Illinois at Urbana Champaign
\vspace{2mm}

To avoid “overlooking” the less frequent classes during training, the concept of focal loss~\cite{lin2017focal} was used for imbalanced datasets. The key idea is to dynamically scale the cross-entropy loss according to the
confidence of the prediction of each class. In addition to the focal loss, the Lovász-Softmax~\cite{berman2018lovasz} function was added, which is shown to be a good surrogate for the intersection-over-union metric used to evaluate the model’s performance~\cite{berman2018lovasz}. Initial tests suggest that using equal weights to combine the focal loss with the Lovász-Softmax loss yields better results.

Two additional input channels were tested and used in the model. The first channel contains the image’s Normalized Difference Vegetation Index (NDVI). The second additional channel used in our work is the image mask. Although pixels outside the valid mask off the image are not considered in the loss function and are not evaluated, they bring relevant information since some classes are spatially correlated with the presence of a non-valid pixel (e.g. waterways are usually marked on the border of the image mask).

The base model used is the ESP Net V2 which is a computationally efficient encoder-decoder~\cite{mehta2018espnet} network. The model was trained from random initialization of its weights, Adam optimizer~\cite{kingma2014adam}. Dropout layers were introduced with a probability of 0.5. The training converged on average in about 35 epochs. The final submission is trained over both training and validation set.

\subsection{Team TJU}
\vspace{-2mm}
\noindent\textbf{Bingchen Zhao, Shaozuo Yu, Siwei Yang, Yin Wang}
\noindent Tongji University
\vspace{1mm}

\begin{figure}[h!]
    \centering
    \includegraphics[width=0.9\linewidth]{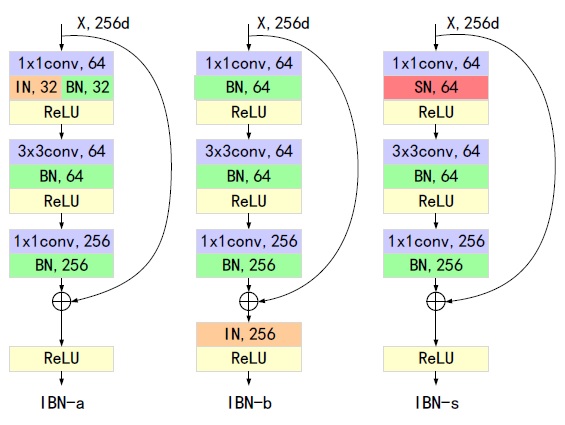}
    \caption{Team TJU: the proposed IBN-s block.}
    \label{fig:tju}
\end{figure}

In the proposed model, switchable normalization~\cite{luo2018differentiable} modules are incorporated with the IBN-Net~\cite{pan2018two} to allow efficient data fusion and reduce feature divergence. Figure~\ref{fig:tju} shows the proposed module. The proposed method aims at resolving the divergence caused by appearance differences between RGB imagery and Near-infrared inputs present in the challenge dataset. In addition, due to potential overlaps of labels in the dataset, the problem is treated as independent binary segmentation tasks for each label type. The Lovász hinge loss~\cite{berman2018lovasz} is used to directly optimize on IoU. Details of this work can be found in our workshop proceedings: \textbf{Reducing the feature divergence of RGB and near-infrared images using Switchable Normalization}.

\subsection{Team Haossr}
\vspace{-2mm}
\noindent\textbf{Hao Sheng, Xiao Chen, Jingyi Su, Ram Rajagopal, Andrew Ng}\\
\noindent Stanford University, Chegg, Inc.
\vspace{1mm}

\begin{figure}[h!]
    \centering
    \includegraphics[width=\linewidth]{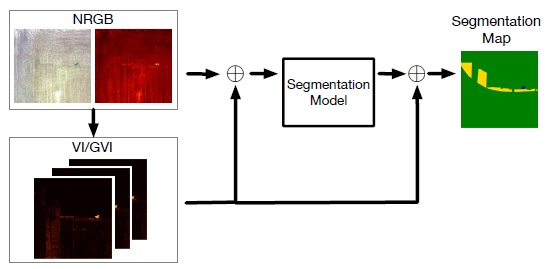}
    \caption{Team Haossr: illustration of the fusion module for the generalized vegetation index.}
    \label{fig:haossr}
    \vspace{-1mm}
\end{figure}

This work focuses on exploring effective fusion techniques for multi-spectral agricultural images. A generalized vegetation index is proposed that is learnable by deep neural networks. The generalized vegetation index module learns a vegetation index feature map given multi-spectral inputs, which can be concatenated with the original color channels and fed into a deep network for inference. In addition, an additive group normalization module is introduced to smoothly train the proposed model with the generalized vegetation index output. An illustration of the fusion module is shown in Figure~\ref{fig:haossr}. Details of this work can be found in our workshop proceedings: \textbf{Effective Data Fusion with Generalized Vegetation Index: Evidence from Land Cover Segmentation in Agriculture}.

\subsection{Team CNUPR\_TH2L}
\vspace{-2mm}
\noindent\textbf{Van Thong Huynh, Soo-Hyung Kim, In-Seop Na}\\
\noindent Chonnam National University, Chosun University
\vspace{1mm}

\begin{figure}[h!]
    \centering
    \includegraphics[width=\linewidth]{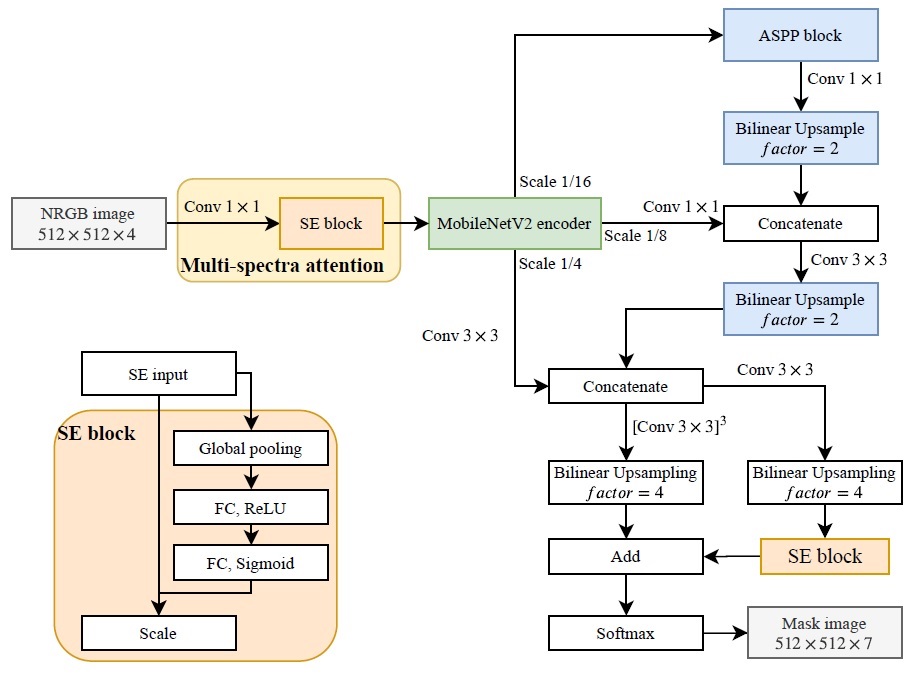}
    \caption{Team CNUPR\_TH2L: pipeline.}
    \label{fig:CNUPR_TH2L}
    \vspace{-1mm}
\end{figure}

A Deep Convolutional Encoder-Decoder architecture is deployed to segment the aerial farmland images. The encoder is based on MobileNetV2~\cite{sandler2018mobilenetv2} with an attention block to assign the contribution of each spectral channel. In the decoder module, ASPP blocks~\cite{chen2016deeplab} are utilized and squeeze-excitation blocks~\cite{hu2018squeeze} are used to upsample the feature map to the original input size. An overview of the method is shown in Figure~\ref{fig:CNUPR_TH2L}.

The network is built with Keras in Tensorflow 2.1 and trained with SGD optimizer. Data augmentation is performed by random flip and/or 90 degree rotation on each image except images that contain only weed clusters. This leads to 38731 images in training set. 6400 images are randomly selected in the training set to optimize the networks in each epoch. A learning rate from 0.05 to 0.3 is used with the cyclical scheduler~\cite{smith2017cyclical}. Due to the highly imbalanced labels in the dataset, class-balanced weighting~\cite{cui2019class} is used with focal loss~\cite{lin2017focal} as objective function. The source code of the method is available at \href{https://github.com/th2l/Agriculture-Vision-Segmentation}{https://github.com/th2l/Agriculture-Vision-Segmentation}.

\subsection{Team TeamTiger}
\vspace{-2mm}
\noindent\textbf{Ujjwal Baid, Shubham Innani, Prasad Dutande, Bhakti Baheti, Sanjay Talbar}\\
\noindent SGGS Institute of Engineering and Technology
\vspace{1mm}

\begin{figure}[h!]
    \centering
    \includegraphics[width=\linewidth]{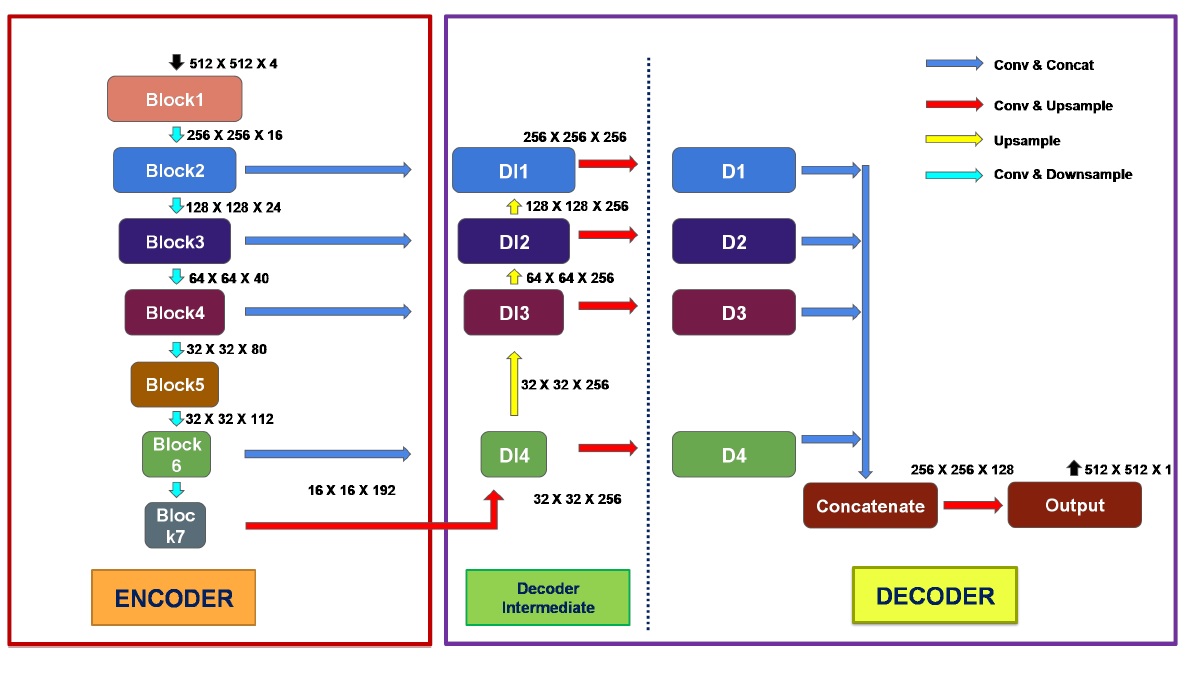}
    \caption{Team TeamTiger: proposed encoder-decoder architecture.}
    \label{fig:ujj}
    \vspace{-1mm}
\end{figure}

The following challenges were incurred for the given segmentation task, (1) Shape and size of the area covered by each anomaly pattern are different; (2) The number of images each class is different; (3) There are overlapping labels. To cope with the challenges mentioned above, an encoder-decoder architecture using EfficientNet~\cite{tan2019efficientnet} and a feature pyramid decoder is used. The proposed encoder-decoder architecture is shown in Figure~\ref{fig:ujj}.

The proposed end-to-end semantic segmentation model is built with Tensorflow 2.0 and Keras. The network is fed with $512\times512\times4$ pixel images with a batch size of four for 100 epochs. To penalize incorrect outputs from the model while training, the Jaccard loss is used with Adam~\cite{kingma2014adam} as the optimizer. The learning rate is kept at 0.001 for initial epochs and then decreased five times whenever the validation does not change for three consecutive epochs.


\section{Conclusion}
To accommodate the rapidly changing computer vision technique in agriculture, the first Agriculture-Vision Challenge targets on efficiently and accurately recognizing several important field patterns from aerial images through semantic segmentation paradigm. Approximately 57 teams around the globe participate in this competition in which 7 leading teams, together with their novel methods, are selected for this paper. Yet our vision of agriculture should be extended beyond segmentation. The inclusive topics about agriculture have initiated many new platforms for future computer vision researches. Therefore, we can expect that, in the near future, more challenging agriculture applications will be brought out, and more powerful computer vision techniques will be developed to better assist these applications as well.

{\small
\bibliographystyle{ieee_fullname}

}

\end{document}